\documentclass[conference]{IEEEtran}
\IEEEoverridecommandlockouts
\usepackage{cite}
\usepackage{amsmath,amssymb,amsfonts}
\usepackage{algorithmic}
\usepackage{graphicx}
\usepackage{textcomp}
\usepackage{wrapfig}
\usepackage{caption}
\usepackage{bbm}
\usepackage{makecell}
\usepackage{array}
\usepackage{multirow}
\usepackage{booktabs}

\makeatletter
\let\NAT@parse\undefined
\makeatother
\usepackage{hyperref} 
\usepackage{cleveref}

\newcolumntype{P}[1]{>{\centering\arraybackslash}p{#1}}
\newcolumntype{R}[1]{>{\centering\arraybackslash}r{#1}}
\newcolumntype{M}[1]{>{\centering\arraybackslash}m{#1}}

\def\BibTeX{{\rm B\kern-.05em{\sc i\kern-.025em b}\kern-.08em
    T\kern-.1667em\lower.7ex\hbox{E}\kern-.125emX}}

\begin{document}

\title{Virtual Fusion with Contrastive Learning for Single Sensor-based Activity Recognition}

\author{\IEEEauthorblockN{Duc-Anh Nguyen}
\IEEEauthorblockA{\textit{University College Dublin} \\
Ireland\\
0000-0001-8613-6854}
\and
\IEEEauthorblockN{Cuong Pham}
\IEEEauthorblockA{\textit{Posts and Telecommunications Institute of Technology} \\
Viet Nam\\
0000-0003-0973-0889}
\and
\IEEEauthorblockN{Nhien-An Le-Khac}
\IEEEauthorblockA{\textit{University College Dublin}\\
Ireland \\
0000-0003-4373-2212}
}

\maketitle

\begin{abstract}
Various types of sensors can be used for Human Activity Recognition (HAR), and each of them has different strengths and weaknesses. Sometimes a single sensor cannot fully observe the user's motions from its perspective, which causes wrong predictions. While sensor fusion provides more information for HAR, it comes with many inherent drawbacks like user privacy and acceptance, costly set-up, operation, and maintenance. To deal with this problem, we propose Virtual Fusion - a new method that takes advantage of unlabeled data from multiple time-synchronized sensors during training, but only needs one sensor for inference. Contrastive learning is adopted to exploit the correlation among sensors. Virtual Fusion gives significantly better accuracy than training with the same single sensor, and in some cases, it even surpasses actual fusion using multiple sensors at test time. We also extend this method to a more general version called Actual Fusion within Virtual Fusion (AFVF), which uses a subset of training sensors during inference. Our method achieves state-of-the-art accuracy and F1-score on UCI-HAR and PAMAP2 benchmark datasets. Implementation is available upon request.
\end{abstract}


\maketitle

\section{Introduction}
\label{sec:introduction}

Human Activity Recognition (HAR) is the task of identifying human activity from sensory data. It has a wide range of applications such as sport performance analysis, exercise monitoring, gaming, human-machine interaction, sign language translation, and health monitoring.

The types of sensors for HAR are also very diverse, each of them has its own advantages and disadvantages. While the use of sensors depends on specific requirements, the two most common ones are camera and wearable sensor. Cameras can capture human-readable data of nearly every movement within its range and context information. Wearable sensors are popular because of their affordability, mobility, and privacy preservation.

With distinct characteristics, each sensor captures different information of the same human activities. In many cases, a single sensor may not be sufficient to recognize activities. For example, methods using only accelerometers are not able to understand the context \cite{Kwolek2016}. Furthermore, \cite{Zhang2020} gave an example that a wrist-worn accelerometer cannot recognize hand activities like wrist turning. Both above papers suggested the use of accelerometer and gyroscope together. Nevertheless, even a wearable device with both accelerometer and gyroscope cannot consider environmental factors. On the other hand, ambient sensors sense everything around them, thereby producing more environmental noise in the data \cite{Hakim2017,Denkovski2022}. Any pet or other people besides the targeted subject would introduce significant disturbance to the system as ambient sensors do not focus solely on the targeted subject \cite{Wang2020}. Note that ambient sensor is a broad term for many specific types of sensors, so they are influenced by different noise sources.

Different sensors can complement each other, and the combination of multiple sensors can result in better accuracy. This has been studied under the name of sensor fusion or multimodal HAR \cite{Aguileta2019,Yadav2021}. However, sensor fusion requires significantly more cost and effort to set up and maintain compared to using a single sensor.

Having identified the limitations of sensor fusion, we propose a novel approach called Virtual Fusion. It aims to take advantage of multiple data modalities available at the training phase, while only using one of them for inference to save deployment, operation, and maintenance costs.

This proposal's ideas focus on contrastive learning where we try to exploit the correlation among data modalities. The hypothesis is that though data come from different sensors, there must be a correlation because they all describe the same user's motion sequence. Although one sensor may contain more useful information than another, deep learning models of both modalities are potentially able to learn from each other what they cannot acquire directly from class labels in a supervised learning setting.

In many situations, the range of sensor choices is broader but still somewhat limited. For instance, access to more than one sensor is available for inference, enabling sensor fusion, and there are more sensors accessible in the training phase. Nevertheless, not all available sensors can be used for model training due to deployment constraints, resulting in a waste of training resources that could be employed to enhance the overall accuracy. To enable more flexible sensor choices for both training and testing and to fully utilize all available training sensors, we extend the idea to infer with a subset of training sensors. For example, multiple wearable sensors and cameras are available for training, but cameras are not permitted in the application environment. In this scenario, we can still utilize all sensors for training and deploy only wearable sensors. This enhances the results while also providing greater flexibility in selecting sensors.

In general, this paper aims at employing various sensors for HAR model training, while only one or a few sensors can be used for inference. Our contributions are as follows:
\begin{itemize}
    \item This paper proposes Virtual Fusion - a new contrastive learning-based method that takes advantage of multiple sensors during training to boost the accuracy of single-sensor inference.
    \item Virtual Fusion is further extended to a more general version for inference with a subset of training sensors, called Actual Fusion within Virtual Fusion (AFVF).
    \item Many experiments are conducted to prove the effectiveness of Virtual Fusion and AFVF, and to compare with other papers. The proposed method achieves state-of-the-art results on benchmark datasets.
\end{itemize}

\section{Related Work}
\subsection{Sensor Fusion}
Multiple sensors can be utilized together in various ways. A survey paper \cite{Aguileta2019} categorized sensor fusion into data level, feature level, and decision level fusion. Meanwhile, the authors of \cite{Yadav2021} divided sensor fusion into 3 categories, namely feature-level fusion, decision-level fusion, and slow fusion. Each name indicates the step in a model at which modalities are combined, while the slow fusion method fuses information at multiple levels throughout the network. Besides that, \cite{Aguileta2019} summarized how modalities are combined (e.g. concatenation at the feature level, voting, averaging at the decision level).

In general, only sensors of different types or positions are fused as they complement each other. \cite{Webber2021} fuses data from accelerometer and gyroscope of the same wearable device. The authors experimented with a range of fusion methods belonging to the three aforementioned categories. \cite{Islam2023} proposed a vision-inertia fusion system, where a different deep network architecture is used for each modality. Besides the commonly used sensors, \cite{Zhu2022} utilizes a sensor network comprised of multiple radars. This system processes raw data into spectrograms and feeds them into a fusion-based HAR model. Likewise, \cite{Cao2023} proposed a method that extracts time-range maps and time-Doppler maps from raw radar signals, then fuses them with a model based on CNN and attention.

There are methods generating new features from the original data, then combining those with the original data. The method in \cite{Pham2021} calculates cosine values of angles between the subject and the floor from human pose data, then concatenates those with the pose data. \cite{Feng2023} fuses radar and camera. To reduce the need for radar data collection and to expand the training set, the authors applied a generative network to generate radar data from image data. \cite{Vu2023} trains a model to extract features from image data that mimic inertial features, then fuses those with features extracted solely from images.

Most of the above approaches need multiple input modalities, and all of them need a model for each modality during both training and inference. Instead, our proposed method only needs those during training. Whereas during inference, it requires no more than a simple single-modality model.

\subsection{Contrastive Learning for HAR}
\label{subsec: contrast for har}
Contrastive learning has been applied to HAR in many ways.
The method in \cite{Hamad2023} applies supervised contrastive learning by contrasting between class labels from the dataset (i.e. maximizing inner class similarity and minimizing outer class similarity) to train the feature extractor. The classifier is then trained separately with a frozen feature extractor.
Cosmo \cite{Ouyang2022} applies sensor fusion and contrastive learning for HAR. Firstly, data of each modality are extracted independently, then combined after some feature projection layers. Then, different augmentations of the fused feature vector are generated and contrasted with each other.

On the other hand, there are also studies contrasting modalities. In ColloSSL \cite{Jain2022}, multiple wearable sensors are contrasted with each other to improve an anchor device's representation, but using the same feature extractor. To ensure data distributions of sensors are akin to each other in the shared model, ColloSSL selects positive and negative sensors in each batch by calculating the maximum mean discrepancy between sensors. Afterward, the model is fine-tuned with a supervised classification task using data from the anchor device. Likewise, COCOA \cite{Deldari2022} contrasts between different sensors but with a separated feature extractor for each, then fine-tunes with a classification task on the fused feature vector. A new loss function is proposed to simultaneously contrast multiple views. \cite{Koo2023} uses 2 distinct feature extractors for accelerometer and gyroscope data. This model is trained with a cross entropy loss for the classification task and a self-supervised learning loss contrasting the 2 types of inertial sensor.

The previous contrastive learning methods require all trained modalities during inference, except for ColloSSL. However, a major drawback is that it only works with modalities of alike distributions, which limits the choice of complementary sensors. Also, the above papers only include the original modalities in contrastive loss computation. Conversely, we suggest that contrasting the fused modality in addition is actually beneficial. More details will be provided to support our claim in \Cref{subsec: contrast fused} and \Cref{subsec: ablation} respectively.

\subsection{More on Pairing in Contrastive Learning}
The above section mentions some contrastive learning papers and what data are contrasted in each. Besides that, the choice of positive and negative pairs is as important. CMC-CMKM \cite{Brinzea2022} contrasts inertia with skeleton. This method finds false negative pairs by calculating intra-modal cosine similarity from the outputs of pre-trained single-modal SimCLR models. Top K most similar pairs are moved from the negative set to the positive set. The drawback is the use of extra separate pre-trained models requiring more computational power. COCOA also contrasts various sensors and its positive pairs are data samples of different sensors and the same timestamp. The distinction is that this method chooses data samples of the same sensor and different timestamps as negative pairs. The effectiveness of these pairing approaches when integrating into our method will be further examined and compared in \Cref{subsec: ablation}.


\section{Virtual Fusion Methodology}
\begin{figure*}[!t]
\captionsetup{justification=centering}
\centerline{\includegraphics[width=5.5in]{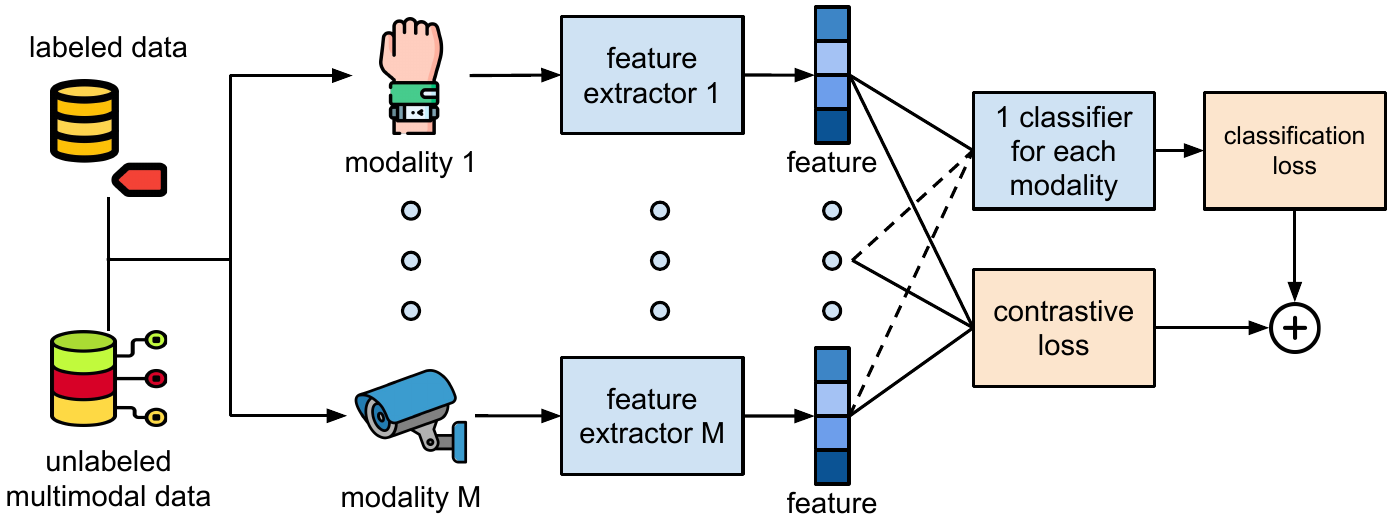}}
\caption{Overall training process of Virtual Fusion. Dotted lines are optional, depending on label availability.}
\label{fig: framework}
\end{figure*}

The proposed method is named Virtual Fusion implying that data of different sensors are trained together to exploit the correlation among them, but they are never directly aggregated for the main task, i.e. classification. This facilitates the use of a single sensor for inference.

\subsection{Problem Definition}
\label{subsec: prob def}
The underlying idea of this study is to learn the correlation between multiple sensing modalities to construct feature maps and aid single-modal classification. It means we have 2 datasets as inputs, one is a labeled dataset and can be either single-modal or multi-modal, and the other is an unlabeled multi-modal dataset. In case the labeled dataset is a multi-modal one, we can train a single-modal classifier for every modality. Data samples are generated using sliding window. Formally, it can be represented as
\begin{equation}
\begin{aligned}
D_{lbl} = \{(x_i^m, y_i) | &i=1,...,N_{lbl}; \\&m\in M_{lbl}; |M_{lbl}| \geq 1\}
\end{aligned}
\end{equation}
with $N_{lbl}$ as the dataset size, $M_{lbl}$ as the set of modalities in the dataset, and $(x,y)$ is a data-label pair. The second dataset is unlabeled and must have at least 2 modalities:
\begin{equation}
\begin{aligned}
D_{ulb} = \{(x_i^m) | &i=1,...,N_{ulb}; \\&m\in M_{ulb}; |M_{ulb}| \geq 2\}.
\end{aligned}
\end{equation}
Unlabeled data is much easier to collect than labeled data. Without having to obtain class labels, we only need to ensure all the sensor devices are time-synchronized. Modalities of $D_{lbl}$ intended for classification should also be in $D_{ulb}$ so they could be trained with contrastive learning (i.e. $M_{lbl} \subset M_{ulb}$), though this is not a strict requirement. At the same time, there could be unlabeled modalities for contrastive learning only.

To train a Virtual Fusion model, we can use either 2 distinct datasets or the same labeled dataset for both $D_{lbl}$ and $D_{ulb}$. We cover both cases in the experiments, but in this section, we use 2 distinct notations when referring to them for readability.

Given these 2 training sets and a test set in the format of $D_{cls}$, we aim to train a classification model for each modality
\begin{equation}
\begin{aligned}
    &f^m: x^m \rightarrow z^m\\
    &c^m: z^m \rightarrow y\\
    &m \in M_{cls}
\end{aligned}
\end{equation}
and evaluate all modalities' classifiers individually on the test set. Here, $f$ is the feature extractor, $z$ is the latent feature vector, and $c$ is the classifier.

\subsection{Contrasting with A Multi-view NT-Xent}
Like CMC \cite{Tian2019}, we calculate a contrastive loss \footnote{In this paper, we use "contrastive loss" as an umbrella term and not a specific loss function.} for every combination of 2 modalities. NT-Xent (the normalized temperature-scaled cross entropy loss) from the SimCLR framework \cite{Chen2020} is utilized as the contrastive loss function between each modality pair in this study.

Given 2 modalities $m_1$ and $m_2$, in every mini-batch of size $B$, we have a set of feature vectors $\{(z^{m_1}_i, z^{m_2}_i) | i=1, ..., B\}$. A mini-batch has $B$ positive pairs, each of them is comprised of 2 samples at the same index $i$ within the batch $(z^{m_1}_i, z^{m_2}_i)$, which also guarantees the same timestamp. The remaining $B-1$ samples are considered negative. The NT-Xent loss is computed for the sample at index $i$ as
\begin{equation}
\label{eqn: ntxent}
    \ell(z^{m_1}_i, z^{m_2}_i) = -\log \frac{\exp(\text{sim}(z^{m_1}_i, z^{m_2}_i) / \tau)}{\sum_{j=1}^B \exp(\text{sim}(z^{m1}_i, z^{m2}_j) / \tau)},
\end{equation}
in which, sim is the cosine similarity function, and $\tau$ is a temperature hyper-parameter.


To support more than 2 modalities, we compute the loss value using \Cref{eqn: ntxent} for all pairs and add them together. The final contrastive loss function for Virtual Fusion is a Multi-view Filtered NT-Xent:
\begin{equation}
    \mathcal{L}_{ctr}(z) = \frac{1}{\binom{M}{2}} \sum_{m_1=1}^{M-1} \sum_{m_2=m_1+1}^M \ell(z^{m_1}, z^{m_2}),
\end{equation}
in which, $M$ is the number of modalities.

While both CMC and SimCLR compute 2-view loss, i.e. $\ell(z^{m_1}, z^{m_2})+\ell(z^{m_2}, z^{m_1})$, we only use 1-view loss as it trains both involved feature extractors. The effectiveness of 2-view loss in our model will be evaluated in \Cref{subsec: ablation}.

\subsection{Positive and Negative Pairs in Contrastive Learning}
The formation of positive and negative pairs is a fundamental part of contrastive learning. In our case, a positive pair comprises data windows of different modalities but of the same timestamp. Therefore, all positive pairs are guaranteed to be correct, as long as sensors are time-synchronized. Negative pairs are sampled randomly, thus, there is a chance that samples in a negative pair are similar. There have been studies proposing to detect and eliminate false negative pairs \cite{Huynh2022,Wang2023}. However, false negative pairs are more problematic if samples of class labels are contrasted with each other. Instead, our method contrasts between samples from different timestamps. Consequently, a false negative pair would only exist if there were 2 identical data samples, which is improbable due to the inconsistency of human motions when repeating the same activity. Even overlapping windows generated by sliding window are not identical because there is always a step between two consecutive windows. We consider partially overlapping windows as hard negatives rather than false negatives. In fact, some papers argued hard negatives are beneficial for contrastive learning \cite{Robinson2021,Deldari2022}.

\subsection{Overall Joint Learning Setting}
Our model is trained with an end-to-end process. All feature extractors and classifiers are trained to minimize a classification loss for each modality and a contrastive loss simultaneously. Thus, this could be considered a joint learning setting. Specifically, average cross entropy loss of all modalities is used as the classification loss
\begin{equation}
    \mathcal{L}_{cls}(y,\hat{y}) = \frac{-1}{M} \sum_{m=1}^M y.\log(\hat{y}^m),
\end{equation}
in which the superscript $m$ denotes the modality index and is not an exponent.

The final loss function to optimize in our proposed method is
\begin{equation}
    \mathcal{L} = \mathcal{L}_{cls} + \mathcal{L}_{ctr}.
\end{equation}

\Cref{fig: framework} shows the overall training process of the Virtual Fusion framework. The wearable sensor and the camera in this figure illustrate that the framework can work with many types of sensors, while in reality, it does not require those specific devices. It can also be expressed as follows:
\begin{equation}
\begin{aligned}
z^m &= f^m(x^m) | \forall m \\
l_{1} &= \mathcal{L}_{cls}(y, c^m(z^m)) | \forall m \in M_{lbl}\\
l_{2} &= \mathcal{L}_{ctr}(z^m) | \forall m \in M_{ulb} \\
loss &= l_1 + l_2.
\end{aligned}
\end{equation}

\subsection{Connecting model components}
\label{subsec: connect components}
Output vectors of feature extractors are used as inputs for both classification and contrastive learning. We want to use the same input vectors for both, so information learned from contrastive loss could directly influence classifiers. Consequently, extractors are made to output the same feature dimension so no projection layer is needed between the feature extractors and the contrastive loss. If the feature dimension must be changed, a projection layer will be considered part of the feature extractor, and its output is used for both class label learning and contrastive learning. Otherwise, if a projector is placed only before contrastive loss, it will lessen the feature extractor's role in optimizing this loss. We also use the same activation function (ReLU) after all feature extractors. Because all features are non-negative after ReLU, cosine similarity values are always between 0 and 1. 

\section{Actual Fusion within Virtual Fusion}
\label{sec: afvf}
Besides single-sensor inference, the proposed method can also work with a subset of training sensors. This means there can be an Actual Fusion within Virtual Fusion (AFVF). For example, we can use 3 sensors for training, but only 2 for inference. These 2 sensors are fused and treated as a unified modality.

\subsection{Early Fusion and Late Fusion}
\begin{figure}[!t]
\captionsetup{justification=centering}
\centerline{\includegraphics[width=0.45\textwidth]{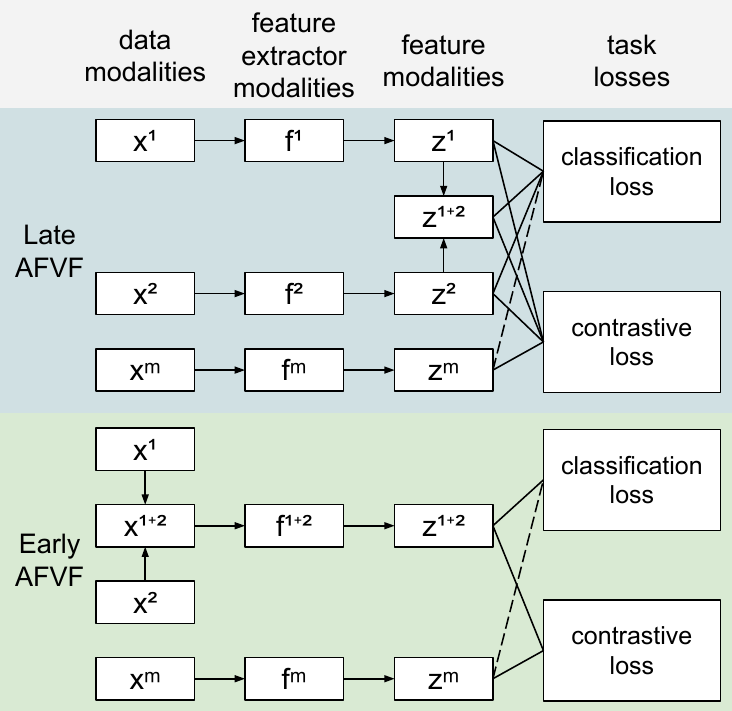}}
\caption{Examples of AFVF that fuses 2 out of multiple modalities. The dotted line connections are only applicable if $m \in M_{lbl}$.}
\label{fig: afvf}
\end{figure}
There are several ways to fuse 2 or more sensors. If the data formats and distributions are the same, they can be fused at either data level or feature level, otherwise, only feature-level fusion is possible. In this paper, we use the terms early fusion and late fusion to refer to data-level fusion and feature-level fusion respectively.

\Cref{fig: afvf} shows examples of AFVF fusing 2 modalities when there are more than 2 in the labeled training set. As illustrated in the figure, late fusion needs more feature extractors than early fusion as it uses 2 extractors to produce the fused feature $z^{1+2}$. As a result, early fusion consumes fewer resources in both the training and the inference phases.

In terms of accuracy, late fusion often performs better as it involves a dedicated feature extractor for each sensor. We also found that late fusion gives better results than early fusion in AFVF. Therefore, we choose late fusion for our AFVF models. In case there is a strict limitation in computational resources, early fusion is a good option.

In early fusion, $f^{1+2}$ controls the number of dimensions of the feature vector $z^{1+2}$. However, in late fusion, features are fused by concatenation, which multiplies the number of dimensions of the fused feature vector. To compute contrastive loss, we must normalize this by adding a fully connected layer as a projector. As mentioned in \Cref{subsec: connect components}, this projector's output will be the input for both contrastive loss and classifier. Suppose that we want to combine $n$ modalities into 1 using late fusion, the fused feature vector will be computed as
\begin{equation}
    z^{fused} = project(concatenate(z^1, ..., z^n)).
\end{equation}

\subsection{Contrast the Fused Modality}
\label{subsec: contrast fused}

\begin{figure}[!t]
\captionsetup{justification=centering}
\centerline{\includegraphics[width=0.45\textwidth]{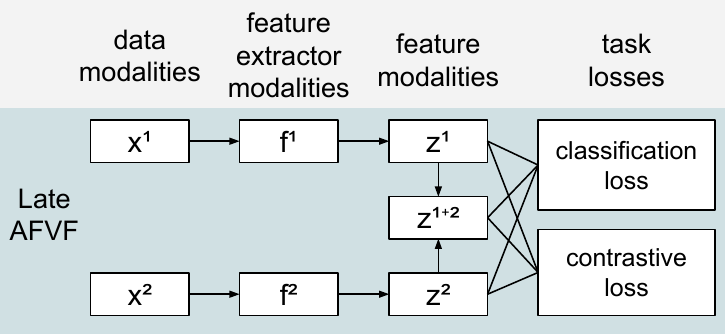}}
\caption{Example of AFVF that fuses all modalities. Early fusion is not applicable.}
\label{fig: afvf fuse all}
\end{figure}
AFVF allows us to infer with any subset of the training sensors. \Cref{fig: afvf} demonstrates that the fused feature vector $z^{1+2}$ is used to compute both loss functions. While the previously reviewed papers only contrast the original modalities with each other, we argue that the fused modality needs to be included in contrastive loss computation as well. As the fused feature is used for classification, this point is actually supported by the point in \Cref{subsec: connect components} that the input features of classifiers and contrastive loss should be the same. Classification performance on the fused modality is improved by training the model to directly contrast it with others.

The source modalities' features $z^1$ and $z^2$ could be excluded from both loss functions because contrasting the fused modality with its sources might seem redundant. Because the fused target feature $z^{1+2}$ is created from the same feature extractors and input data of the source ones, the model tends to retain some information from the source features in the target one for easier contrasting and matching. However, we still keep source modalities for loss computation because it also means the model is trained to fuse multiple input streams while retaining their descriptive features. Also, this does not require more computational power as the number of feature extractors remains the same.

If we want to use all training sensors for testing, late AFVF comes in handy as illustrated in \Cref{fig: afvf fuse all}. Early AFVF is not applicable 
as it does not produce features for the source modalities.

\section{Experiments}
This section describes our experiment settings, experiments that validate our method, and comparisons with other studies on benchmark datasets\footnote{Implementation is available upon request to nda.3157@gmail.com or duc-anh.nguyen@ucdconnect.ie.}.

As presented in the previous sections, Virtual Fusion models do not involve any actual fusion and utilize a single modality in testing. While AFVF models actually combine some modalities and utilize a subset of training sensors for testing. These terms are also used when reporting experiments.

\subsection{Deep Model and Configuration}
We use ResNet 1D \cite{resnet1d_github} as the feature extractor for all experiments. All models are trained until the validation F1-score does not increase in 30 consecutive epochs. The best model checkpoint on the validation set is then used for testing. We use Adam optimizer with a learning rate of $10^{-3}$, which is divided by 10 after 15 consecutive epochs that the validation score does not improve. The batch size is 32. When training Virtual Fusion models, half of the batch is labeled data, and the other half is unlabeled data. Labeled data are resampled so that all classes are trained with the same frequency. The reported F1-scores are averages of 3 runs.

\subsection{Experimental Validation}

\begin{table*}
\centering
\caption{Public HAR datasets for experimental validation}
\label{tab: baseline datasets}
\setlength{\tabcolsep}{3pt}
\begin{tabular}{@{}llccM{65pt}M{65pt}M{65pt}@{}}
\toprule
\textbf{Dataset} & \textbf{Used modalities} & \textbf{No. labels} & \textbf{No. subjects} & \textbf{Train subjects} & \textbf{Valid subjects} & \textbf{Test subjects}\\ \midrule
CMDFall\cite{Tran2018} & \makecell[l]{wrist accelerometer,\\waist accelerometer,\\3D skeleton} & 20 & 50 & odd subject IDs & subject IDs divisible by 10 & even subject IDs not divisible by 10\\ \hline
UP-Fall\cite{Lourdes2019} & \makecell[l]{wrist accelerometer,\\waist accelerometer,\\2D skeleton extracted from\\image with OpenPose\cite{Cao2018}} & 11 & 17 & odd subject IDs not divisible by 5 & subject IDs divisible by 5 & even subject IDs not divisible by 5\\ \midrule
FallAllD\cite{Saleh2021} & \makecell[l]{wrist accelerometer,\\waist accelerometer} & 2 & 12 (out of 15) & 1, 3, 12 & 4, 5 & 2, 9, 10, 11, 13, 14, 15\\ \bottomrule

\multicolumn{7}{p{430pt}}{Only sensors and class labels we use are listed in the table. For FallAllD, we only use data from 12/15 subjects as not all wore both sensors. Subject ID starts from 1.}
\end{tabular}
\end{table*}

This section reports the experiments that validate the improvements achieved by our method and the design of its components.

\subsubsection{Data Preparation}
Datasets used for experiments are listed in \Cref{tab: baseline datasets}. Each dataset is divided by subjects into training, validation, and test sets. For FallAllD, we allocate less data in the training set to make it more challenging because there are only 2 classes. The subjects in FallAllD are divided to achieve comparable class ratios in three sets. Results are reported as macro F1-scores for CMDFall and UP-Fall, and binary F1-scores for FallAllD with fall as the positive class and non-fall as the negative class.

Besides the labeled data, we need an unlabeled multimodal training set. CMDFall is chosen for this purpose because it is a continuous dataset and subjects could move freely with little scenario restriction. Also, it has 3D skeleton data, which is very informative to learn correlation among modalities. Except for the case of 2D skeleton in the UP-Fall dataset, we use UP-Fall itself as the unlabeled data because CMDFall does not have this modality. All unlabeled datasets are also split according to \Cref{tab: baseline datasets} to prevent data leakage.

All data are converted to the same sampling rate of 50Hz for accelerometer and 20Hz for skeleton. Skeleton data are normalized by moving the skeleton to the coordinate system's origin to remove differences in the subject's relative position. 2D skeleton is further normalized by dividing joint coordinates by skeleton size to remove differences in subject-camera distance. Finally, sliding window with a 4-second window size is run to obtain data samples.

\subsubsection{Augmentation}
In general, we aim to choose augmentations that preserve the semantic meaning of the data, especially when an unsuitable augmentation might change it and contaminate contrastive learning. Based on the data types used in our experiments, we apply augmentations as follows:
\begin{itemize}
    \item Accelerometer: 3D rotation with random axis.
    \item 2D skeleton: Horizontal flip.
    \item 3D skeleton: Rotation around the Z-axis.
\end{itemize}
For rotation, a random angle is drawn within a pre-defined range every time a data sample is queried for training. For the classification task, we tune this range to achieve the best result on each dataset. For the contrastive learning task, we set this range to its maximum, which is [-180, 180] degrees. Stronger augmentation for contrastive learning than supervised learning was also mentioned in the SimCLR paper \cite{Chen2020}.

\subsubsection{Comparison With Baselines}
\label{subsec: exp baseline}

\begin{table}
\caption{F1-score comparison of the proposed method with the baselines}
\label{tab: exp baseline}
\centering
\setlength{\tabcolsep}{5pt}
\begin{tabular}{@{}llrrr@{}}
\toprule
\textbf{Dataset} & \textbf{Modality} & \textbf{\makecell{Single-\\sensor}} & \textbf{\makecell{Fusion\\(all sensors)}} & \textbf{\makecell{Virtual\\Fusion}} \\ \midrule
\multirow{4}{*}{CMDFall} & waist & 0.6734 & \multirow{3}{*}{0.7670} & 0.7027 \\
& wrist & 0.5043 &  & 0.5553 \\
& skeleton & 0.6739 &  & 0.7394 \\ \cmidrule{2-5}
& waist+wrist & 0.7223 &  & 0.7455 \\ \midrule
\multirow{4}{*}{UPFall} & waist & 0.7227 & \multirow{3}{*}{0.8777} & 0.7604 \\
& wrist & 0.5007 &  & 0.5840 \\
& skeleton & 0.8125 &  & 0.8449 \\ \cmidrule{2-5}
& waist+wrist & 0.6880 &  & 0.7635 \\ \midrule
\multirow{3}{*}{FallAllD} & waist & 0.9286 & \multirow{2}{*}{0.9277} & 0.9452 \\
& wrist & 0.8863 &  & 0.9229 \\ \cmidrule{2-5}
& waist+wrist & 0.9277 &  & 0.9587 \\ \bottomrule
\multicolumn{5}{l}{\makecell[l]{Macro F1 for CMDFall and UP-Fall,\\binary F1 for FallAllD.}}
\end{tabular}
\end{table}

This section compares Virtual Fusion with single-modal and fusion models. Besides single-sensor inference for Virtual Fusion, we also experiment with AFVF using 2 accelerometers and omitting skeleton while testing. This simulates the scenario where cameras are not allowed. Details are presented in \Cref{tab: exp baseline}, where the AFVF results are placed in the last row of each dataset.

Virtual Fusion outperforms single-modal in all comparisons while using the same sensor for testing. The most remarkable improvement is in the UP-Fall dataset's waist accelerometer, where the gap is more than 8\%. Virtual Fusion even exceeds the fusion model in the FallAllD dataset. In the other 2 datasets, its scores come very close to the score of actual fusion.

Fusion generally improves or at least retains the results compared to using a single sensor, except for the case of the UP-Fall dataset. Adding the wrist accelerometer worsens the results of fusion. This implies the sensor contains much noise and only confuses the model. Nevertheless, AFVF with waist and wrist accelerometers still works and improves the result.

\subsubsection{Ablation Study}
\label{subsec: ablation}

\begin{table}
\caption{Influence of some model components}
\label{tab: ablation}
\centering
\begin{tabular}{@{}lrrr@{}}
\toprule
\textbf{Model variants} & \textbf{CMDFall} & \textbf{UP-Fall} & \textbf{FallAllD} \\ \midrule
2-view loss & 0.7227 & \underline{0.7548} & \underline{0.9560} \\
CMC-CMKM pairing & 0.7368 & 0.7391 & 0.9515 \\
COCOA loss function & 0.7161 & 0.7130 & 0.9379 \\
exclude fused modal & 0.7383 & 0.7414 & 0.9545 \\
exclude original modals & 0.7055 & 0.7311 & 0.9297 \\
early AFVF & \underline{0.7396} & 0.7466 & 0.9473 \\
\textbf{late AFVF (proposed version)} & \textbf{0.7455} & \textbf{0.7635} & \textbf{0.9587} \\ \bottomrule
\multicolumn{4}{p{210pt}}{The best and second-best F1-scores are highlighted in bold text and underline respectively.}
\end{tabular}
\end{table}

We try replacing or removing some components of Virtual Fusion to see how it affects the results. For succinctness, we only choose one modality to test in this section. Also, because this experiment involves AFVF, we choose the combination of waist and wrist accelerometers as the modality for this ablation study. Specifically, we examine the following modifications to the proposed framework one at a time:
\begin{itemize}
\item Replacing 1-view contrastive loss with 2-view,
\item Integrate the positive mining and negative pruning approach in CMC-CMKM \cite{Brinzea2022} into the training process,
\item Using COCOA \cite{Deldari2022} as the contrastive loss function,
\item Using early AFVF instead of late AFVF,
\item Excluding the fused modality in contrastive loss,
\item Excluding the original modalities in both loss functions,
\end{itemize}
where the last 2 modifications have been emphasized in \Cref{subsec: contrast fused}. \Cref{tab: ablation} shows that the proposed version surpasses other model variants, thus it further proves the capability of our model components.

\subsection{Comparison With Other Studies}
\label{sec: Comparison With Other Studies}
In this section, we compare the proposed method with other recent studies on benchmark datasets.

\subsubsection{Benchmark Datasets}
For fair comparisons, we only use datasets that clearly define the training set and the test set, or are split in the same way by many other papers. Accordingly, we can only make comparisons with papers using the same deterministic split. Also, because our method needs multiple sensors in the training set, we only search for multimodal datasets so that no external training data is needed. With these criteria, this section uses the below datasets.

\textbf{UCI-HAR} \cite{Anguita2013} is a popular benchmark dataset for HAR. It includes 6 class labels and 30 subjects. It has already been clearly formatted and is ready to use. The sampling rate of both accelerometer and gyroscope is 50Hz. All data have already been divided into windows of 2.56 seconds. UCI-HAR is partitioned into a training set and a test set, containing 7352 and 2947 windows respectively.

\textbf{PAMAP2} \cite{Reiss2012} has 9 subjects, 12 protocol and 6 optional activity class labels. Only 12 protocol activities are used for evaluation. It does not provide a pre-partitioned training set and test set or a fixed window size. However, most papers using this dataset used data from subjects 5 and 6 for testing and the rest for training. We follow this split for evaluation. We use a window size of 5.12 seconds and 50\% overlapping. For the classification task, data of transient activities are discarded as noted by the dataset authors. For the contrastive learning task, we keep these data as it does not involve class labels.

Unlike the previous experiments, both datasets in this section do not have data from cameras.

\subsubsection{Augmentation}
We employ several additional augmentation techniques from \cite{Um2017}. When training on the UCI-HAR dataset, we apply scale augmentation for the classification task. As for the contrastive learning task, we add 2 more augmentations, which are 3D rotation and time warping. For PAMAP2, magnitude warping and time warping are applied to both tasks but with a stronger intensity for the contrastive learning task.

\subsubsection{AFVF}
While AFVF is applied to both datasets, sensors used for classification and contrastive learning differ.

The UCI-HAR dataset has an accelerometer and a gyroscope on the same device. These 2 modalities are fused for both classification and contrastive learning. The 2 original modalities are also used for contrastive loss computation. This is exactly like the example in \Cref{fig: afvf fuse all} where the final classification result is from the fused modality.

The PAMAP2 dataset has a heart rate sensor and 3 IMU devices worn on the chest, hand, and ankle of every subject. Each IMU device has several types of built-in sensors. We use acceleration, gyroscope, magnetometer, and orientation data from all 3 devices. In AFVF, late fusion is applied for classification modalities. To reduce the number of feature extractors needed, early fusion is applied for the other modalities. In summary, there are 9 modalities for loss computation: 3 accelerometers, 3 magnetometers, late fusion of all accelerometers and magnetometers, early fusion of all gyroscopes, and early fusion of all orientations. The final classification result is from the late fusion one.

\subsubsection{Score Comparison}
\begin{table}
\centering
\caption{Comparison on benchmark datasets}
\label{tab: benchmark compare}
\setlength{\tabcolsep}{3pt}
\begin{tabular}{@{}lrrrr@{}}
\toprule
\multirow{2}{*}{\textbf{Method}} & \multicolumn{2}{c}{\textbf{UCI-HAR}} & \multicolumn{2}{c}{\textbf{PAMAP2}} \\
& \textbf{accuracy} & \textbf{macro F1} & \textbf{accuracy} & \textbf{macro F1} \\ \midrule
Real time CNN \cite{Ignatov2018} & 0.9763 & 0.9762 &  &  \\
Layer-wise CNN \cite{Teng2020} & 0.9698 & 0.9697 &  &  \\
DanHAR \cite{Gao2021} &  &  & 0.9316 &  \\
Ensem-HAR \cite{Bhattacharya2022} & 0.9505 &  &  &  \\
CNN and AOA \cite{Dahou2022} & 0.9523 & 0.9533 &  &  \\
Marine predators \cite{Helmi2023} &  &  &  & 0.9276 \\
Multi-ResAtt \cite{Mohammed2023} &  &  & 0.9319 & 0.9296 \\
DCapsNet \cite{Sezavar2024} & \underline{0.9843} &  &  &  \\
Contrastive Distillation \cite{Xu2023} & 0.9657 & 0.9656 &  &  \\
Contrastive Supervision \cite{Cheng2023} &  &  & \underline{0.9322} & \underline{0.9297} \\
\textbf{AFVF} & \textbf{0.9861} & \textbf{0.9865} & \textbf{0.9672} & \textbf{0.9665} \\ \bottomrule
\multicolumn{5}{p{220pt}}{The best and second-best F1-scores are highlighted in bold text and underline respectively.}
\end{tabular}
\end{table}

\Cref{tab: benchmark compare} shows the evaluation results of other papers from 2018 to 2023 in comparison with our results. Some papers focused on deep model architecture, and some focused on learning strategy. AFVF achieves the highest scores on both benchmark datasets. Note that our method is model agnostic, so it could be used with any other deep models to further improve the results.

\section{Discussion and Conclusion}
In this paper, we propose Virtual Fusion to take advantage of unlabeled multimodal data for training, while only using one sensor for inference. We also extend it to a more general version called AFVF for inference using only a subset of training sensors. Our method facilitates flexibility in choosing sensors for HAR applications.

While it is good to have labeled data, Virtual Fusion can also exploit unlabeled multimodal data for representation learning. This is potential because collecting more unlabeled data is easier and considerably cheaper than labeled data.

We show that the input feature vectors of contrastive loss must be the same as those of the classifiers to directly support the classification task. This also means the fused modality should be included in the contrastive loss function, and not only the original ones. Note that including the fused feature does not imply excluding the original ones.

The experiments show that Virtual Fusion outperforms single-sensor training, and in some cases, it even surpasses actual sensor fusion. Also, AFVF achieves SOTA accuracy on benchmark datasets.

Because multiple datasets can be used together for training, the employment of domain adaptation or generalization techniques could be a potential direction. Meanwhile, the fact that each dataset may have a different set of class labels is challenging for these techniques.

Another potential future work is to investigate the effects of the number of sensors and sensor characteristics on Virtual Fusion. This seeks to answer what kinds of sensors are complementary, and whether more sensors result in better accuracy. This is important because using more sensors requires more data collection and computational resources.


\bibliographystyle{IEEEtran}
\bibliography{main}

\end{document}